\documentclass[10pt,twocolumn,letterpaper]{article}

\newcommand{\Tref}[1]{Table~\ref{#1}}

\newcommand{\Fref}[1]{Fig.~\ref{#1}}
\newcommand{\Sref}[1]{Sec.~\ref{#1}}

\newcommand{\etal}{\textit{et al. }}

\usepackage{iccv}
\usepackage{times}
\usepackage{epsfig}
\usepackage{graphicx}
\usepackage{amsmath}
\usepackage{amssymb}
\usepackage{algorithm}
\usepackage[noend]{algpseudocode}
\usepackage[english]{babel}
\usepackage[T1]{fontenc}
\usepackage[utf8]{inputenc}
\usepackage{booktabs} 
\usepackage{colortbl}

\usepackage[english]{babel}
\usepackage[T1]{fontenc}
\usepackage[utf8]{inputenc}
\usepackage{booktabs} 

\usepackage{subfigure}
\usepackage{adjustbox}
\usepackage{booktabs}
\iccvfinalcopy 
\usepackage{hyperref}

\begin{document}
\ificcvfinal\pagestyle{empty}\fi

	\title{Pixel-Level Matching for Video Object Segmentation using Convolutional Neural Networks}
\author{
	Jae Shin Yoon$^\dagger$ $^\ddagger$
	\hspace{4mm}Francois Rameau$^\ddagger$ 
	\hspace{4mm}Junsik Kim$^\ddagger$
	\hspace{4mm}Seokju Lee$^\ddagger$
	\hspace{4mm}Seunghak Shin$^\ddagger$
	\hspace{4mm}In So Kweon$^\ddagger$
	\vspace{2mm} 
	\\ 
	$^\dagger$UMN, Minneapolis, MN
	\hspace{4mm}
	$^\ddagger$KAIST, South Korea
	\\
	\vspace{3mm} 
	{\tt\small yoon0074@umn.edu, \{frameau, Jskim2, sjlee, shshin\}@rcv.kaist.ac.kr, iskweon@kaist.ac.kr}
	\\
}

	\maketitle
	\thispagestyle{empty}
%



\begin{abstract}
	We propose a novel video object segmentation algorithm based on pixel-level matching using Convolutional Neural Networks (CNN). Our network aims to distinguish the target area from the background on the basis of the pixel-level similarity between two object units. The proposed network represents a target object using features from different depth layers in order to take advantage of both the spatial details and the category-level semantic information. Furthermore, we propose a feature \textit{compression} technique that drastically reduces the memory requirements while maintaining the capability of feature representation. Two-stage training (pre-training and fine-tuning) allows our network to handle any target object regardless of its category (even if the object's type does not belong to the pre-training data) or of variations in its appearance through a video sequence. Experiments on large datasets demonstrate the effectiveness of our model - against related methods - in terms of accuracy, speed, and stability. Finally, we introduce the transferability of our network to different domains, such as the infrared data domain.
\end{abstract}
\vspace{-1mm}

%

\vspace{-3.5mm}
\section{Introduction \& Related Works}
\vspace{-1mm}
Video object segmentation refers to the propagation of the mask of an initial object(s) from the first to the last frame of a video sequence. With it, users can determine the pixel-level foreground masks of every image from single key frame supervision. Separating the foreground from the background in a video is a fundamental problem with high applicability to many video based tasks including video summarization, stabilization, retrieval, and scene understanding. For these reasons, video object segmentation has been studied intensively, but it still demonstrates poor results in real world scenarios.

\begin{figure}[t]
	\centering
	\begin{tabular}{@{}c}
		\hspace{-2mm}{\includegraphics[width=1\linewidth]{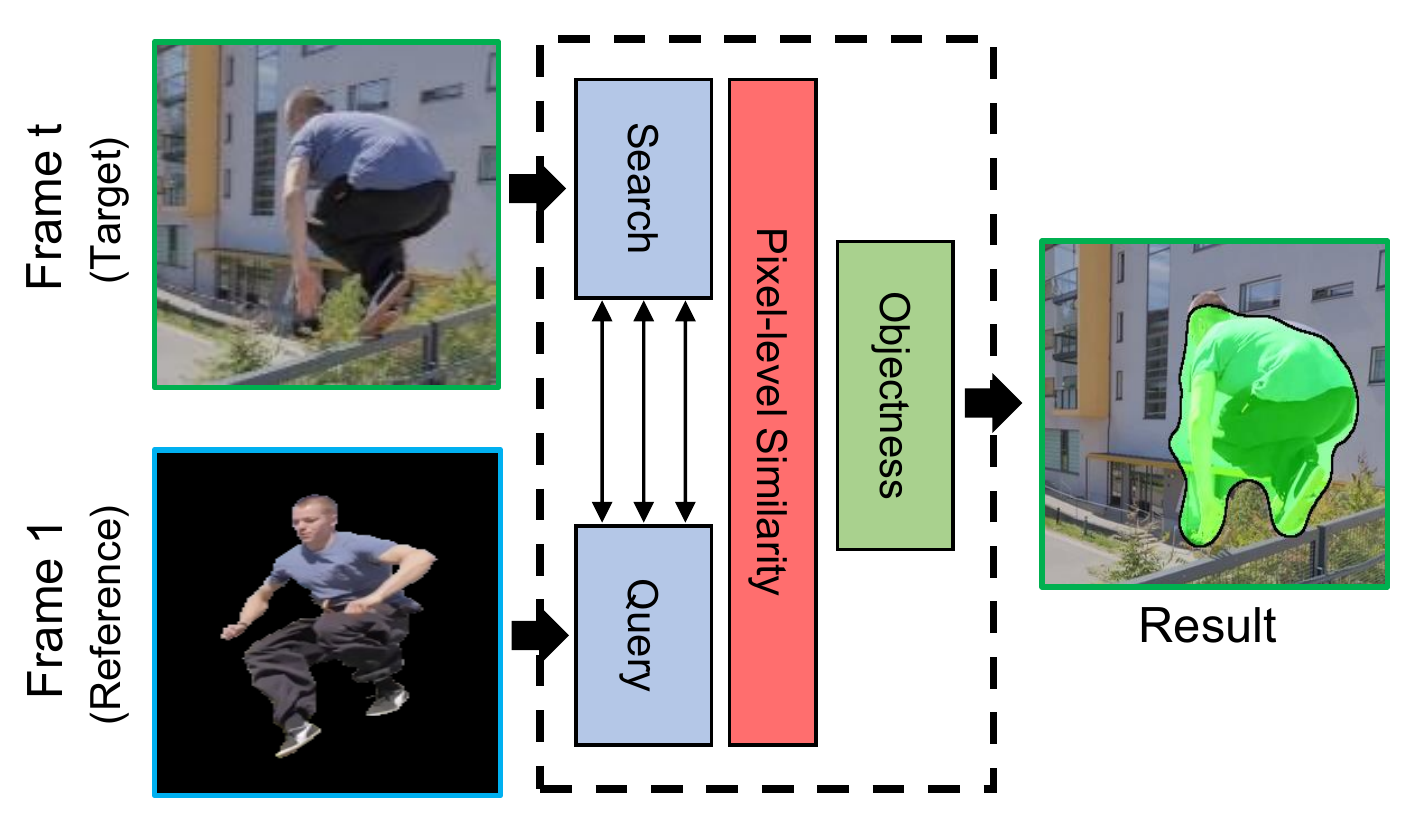}}\hspace{1mm}
	\end{tabular}
	\caption{Example of a result of the proposed pixel-level matching network. Here, a completely segmented reference frame is fixed to the query and a sample target frame is fed to the search input. Each colored box is associated with the same color in \Fref{fig:network}.}
	\label{fig:tiger}
	\vspace{-3mm}
\end{figure}

Most recent approaches \cite{dondera2014interactive,price2009livecut,wang2005interactive,badrinarayanan2010label,ramakanth2014seamseg,Tsai_CVPR_2016} separate discriminative objects from a background by optimizing an energy equation under various pixel graph relationships. For instance, fully connected graphs have been proposed in \cite{perazzi2015fully} to construct a long range spatio-temporal graph structure robust to challenging situations such as occlusion. In another study~\cite{jain2014supervoxel}, the higher potential term in a supervoxel cluster unit was used to enforce the steadiness of a graph structure. More recently, non-local graph connections were effectively approximated in the bilateral space~\cite{marki2016bilateral}, which drastically improved the accuracy of segmentation. However, many recent methods are too computationally expensive to deal with long video sequences. They are also greatly affected by cluttered backgrounds, resulting in a drifting effect. Furthermore, many challenges remain partly unsolved, such as large scale variations and dynamic appearance changes. The main reason behind these failure cases is likely poor target appearance representations which do not encompass any semantic level information. 

Recently, Convolutional Neural Networks (CNN) have been extensively applied to various vision tasks given their ability to encapsulate semantic information in the object representation task itself. In spite of this advantage, however, only a few attempts \cite{Cae+17,Perazzi2017,jampani:vpn:2017} have been made to solve video object segmentation problems using a CNN. Three major causes can be considered: The lack of a training dataset is the primary problem. Indeed, CNN generally requires training with abundant and representative datasets for specific classes. However, it is difficult to account for all types of target object classes given that the class type is always assumed to be undefined for video object segmentation purposes.
Many recent benchmarks \cite{fan2015jumpcut,FliICCV2013,perazzibenchmark} (including pixel-level ground-truth labeling) provide a partial solution to this problem but nonetheless cannot cover all possible classes. Secondly, training a network exclusively with the initial frame of a video usually leads to over-fitting. If the network is over-fitted to a specific object appearance in the first image, it cannot deal with appearance changes of the target object in the subsequent frames. Finally, the localization of the target object using a CNN is also a challenging issue, especially in relation to pixel-level accuracy. Most CNN structures encode category-level semantic information using features from deep layers generally exploited for object classification. However, they cannot preserve the spatial details of the target object, which are the key components for object localization. For this reason, the idea of combining higher-level features with lower-level ones has attracted much attention in relation to object localization but a unified and elegant method has yet to be proposed.

\begin{figure*}[t]
	\centering
	\begin{tabular}{@{}c}
		{\includegraphics[width=1\linewidth]{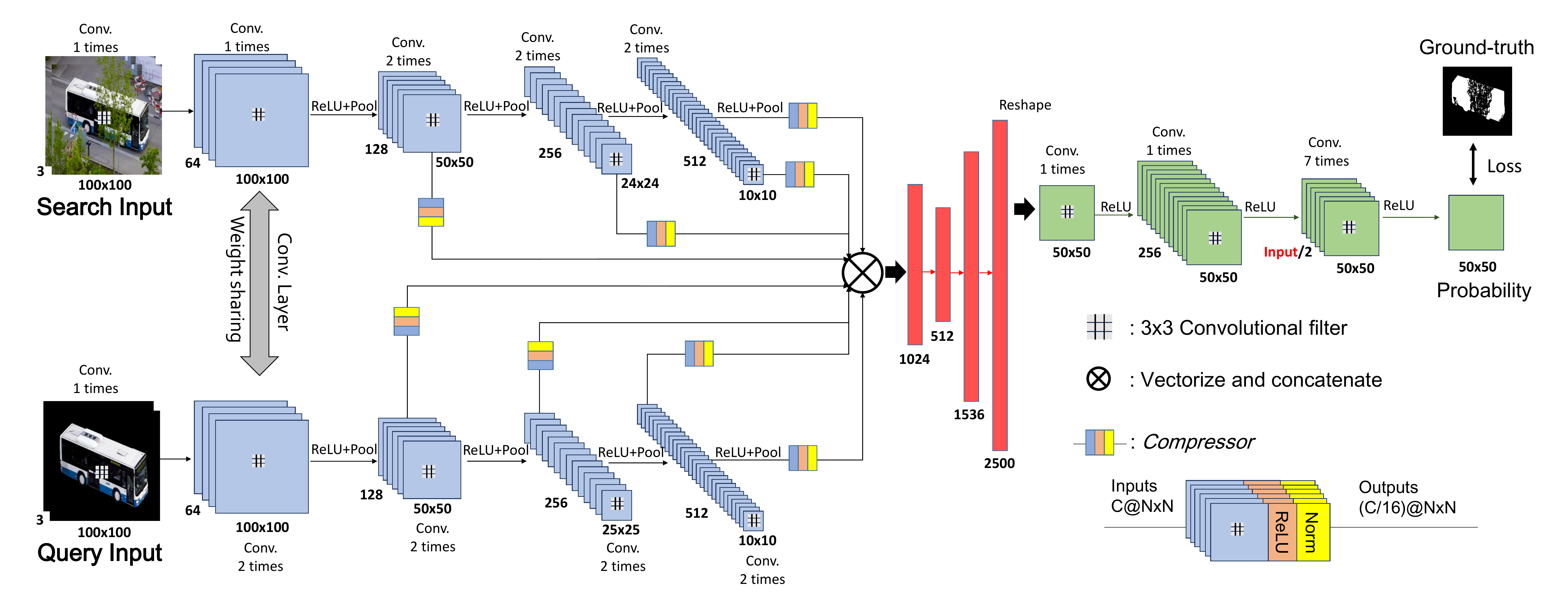}}\hspace{0.1mm}
	\end{tabular}
	\vspace{-4mm}
	\caption{The architecture of the proposed pixel-level matching network. Multi-layered features are extracted from a Siamese structure (blue). Here, all convolutional layers share their weights with mirrored layers including \textit{compressors}. The pixel-level similarity is then encoded via three more hidden fully connected layers (red). Finally, we discriminatively enforce the object coherency through multiple usages of convolutional layers, finally classifying each pixel into the background and the foreground (green). Zero-padding is properly used to fit the output size, while the pooling size is 2$\times$2 with a stride of 2. 
	}
	\label{fig:network}
	\vspace{-2mm}
\end{figure*}
In order to solve the aforementioned problems, our method is mainly inspired by recent breakthroughs in visual tracking. In a sense, visual tracking shares many common features with video object segmentation (i.e. propagating the initial object mask through a series of images). CNN based visual tracking is also a relatively new trend, but various approaches have already been proposed. We can distinguish two types of visual tracking methods using a CNN, as described below:\\ 
1) \emph{Generative model} based tracking aims statistically to describe the appearance of a target and to track the bounding box with the best score compared to previous ones. Some works~\cite{ma2015hierarchical,qi2016hedged} exploit hierarchical features from different layers to create robust correlation filters. However, these approaches cannot integrate this idea into an end-to-end CNN structure. Nam \etal~\cite{nam2015learning} pre-train a network using tracking datasets and fine-tune the model using the initial appearance of the target (in the first image), such that their tracker can be adapted to any type of object. Tao \etal~\cite{tao2016siamese} validated that matching with Siamese structure can be robust to various challenging scenarios without model updates.\\
2) \emph{Discriminative tracking} involves the learning of a model that separates distinguishable target objects from the surrounding background. Wang \etal~\cite{wang2015visual} effectively deal with the identification problems using the features from conv4 and conv5 layers simultaneously. The criterion used to select the layers, however, causes confusion to the network. In other work~\cite{wang2015transferring}, a pre-trained network is frequently updated with the appearance of the object to track. For this purpose, the authors reshape the last fully connected layer encoded with the target objectness and calculate the element-wise loss with a conventional regression model.\\
In consideration of the above tracking approaches, the proposed network embodies the features from lower to higher levels in an end-to-end process. Our learning method also consists of two steps to propose an arbitrary target class adaptation and to cope with appearance variations. The feature extraction part of our network is designed with a Siamese structure to improve the robustness against challenging scenarios. Using the proposed network, our video object segmentation strategy consists of the matching of an object of interest (initialized in the first frame) with subsequent images until the end of a sequence. Therefore, the proposed network must be trained to perform semantic matching between non-successive frames which contain target appearance variations as shown in \Fref{fig:tiger}.

Overall descriptions of the proposed network and the target object mask propagation strategy are described in \Fref{fig:network} and \Fref{fig:inference} respectively. The main contributions of this paper are three folded as presented below:	

$\bullet$ We propose a novel pixel-level matching network for video object segmentation. Our method shows good computational efficiency as well as higher performance capabilities compared to previous graph based approaches. $\bullet$ We propose a feature \textit{compression} technique that drastically reduces the memory requirements of the network but that also maintains its representation ability. $\bullet$ We experimentally validate the transferability of our network pre-trained on RGB data wiith different domain like infrared data.

\vspace{-2mm}
\section{Proposed Method}\label{sec:proposed}
\vspace{-1mm}

If we directly train the proposed network from scratch (random weight parameters) using only the initial frame of a new sequence, the network weights are over-fitted to the first frame object and cannot handle appearance changes. To prevent this, we split the training into two stages. The first involves network pre-training (off-line). In this stage, training involves semantic matching inference in order to deal with appearance variations using a dataset consisting of 300,000 image pairs. The second stage is network fine-tuning (on-line) using the appearance of the object in the first image. This step is indispensable because the target object does not necessarily belong to any pre-trained object class. This makes our network versatile enough to deal with any arbitrary target.  

In light of this two-stage learning process, we present specific methods by which to train our pixel-level matching network including its architecture details. We then explain how our model can be applied directly to video object segmentation tasks.

\subsection{Pretraining Pixel-Level Matching Network}\label{sec:pretraining}
\vspace{-2mm}
The proposed network is composed of two main parts: pixel-level similarity encoding and target objectness decoding. Our model is pre-trained using 30 video sequences from the~\textit{DAVIS}~\cite{perazzibenchmark} dataset which contains various challenging scenarios. A global description of our network is available in \Fref{fig:network}.


\textbf{Network inputs}\ \ \
Since the first part of our network is a Siamese structure, it requires two types of inputs: a reference for the query stream and a target for the search stream. To make our network robust to dynamic appearance variations during the learning process, we choose the target and reference frames randomly (in the same sequence) in order to avoid successive frames (with high similarity).

To generate the query stream dataset, we randomly select 20 reference frames in each training video. For each frame, we crop the bounding box containing the reference object leaving margins around it (25\% larger than the original object box size). This region of interest is then completely segmented using the ground-truth label and resized into a 100$\times$100p image patch.

Regarding the search stream data, we randomly select six target frames associated with each reference frame in the same sequence. We then generate the search stream data by cropping and resizing (into 100$\times$100p) multiple bounding boxes at various locations around the target object with three different margins sizes ($\rho_{1}$, $\rho_{2}$ and $\rho_{3}$). By doing this, we can train the model to deal with localization and scale variation. The generated search stream data is further augmented by flipping and rotating. Here, the target object is always fully or partially included in the cropped box (overlapping minimum: 50\%). In our experiments, training with hard negative data which does not contain any part of a target object has proved to be rather inefficient.

\textbf{Pixel-level matching}\ \ \
To encode the pixel-wise similarity between the search and query processes, our network initially extracts the features from inputs through several combinations of convolution, max pooling and Rectified Linear Units (ReLU). In the field of visual tracking, one of the major issues is to determine the optimal number of layers to ensure an effective representation of a target object. It was recently found that too many deep layers can be redundant because visual tracking is a binary classification task (foreground or background). Therefore, many works~\cite{tao2016siamese,wang2015visual,ma2015hierarchical,qi2016hedged} have exploited VGGNet~\cite{simonyan2014very} or proposed much lighter structures~\cite{wang2015transferring,nam2015learning} which demonstrate high performance in terms of accuracy and processing time. Given that our task also belongs to binary classification, we designed our feature extractor similarly. \Fref{fig:network}-(blue) shows our feature extraction part. Because our model starts with a Siamese structure, the weights for feature extraction are shared between the search and query streams.

In the fully connected (FC) layers (\Fref{fig:network}-(red)), the similarity between the query and search inputs is globally encoded using the extracted features. Here, directly feeding the output features from the last layer in the Siamese structure to the initial FC layer is a good strategy to generate the semantic information of a target object. It can, however, also cause critical localization and identification issues due to the lack of spatial details. On the other hand, the use of lower features alone makes it easy to lose the semantic information. To solve this dilemma, we exploit all of the feature instances from the lower to the higher layers and stack them on the initial FC layer. We then use three more hidden FC layers to globally combine the multi-level feature instances which encode the similarity between two object units. The size and number of FC layers are similar to that in earlier work~\cite{wang2015transferring}, which reduces the strong correlation between neighboring pixels to prevent over-fitting problems. A 50$\times$50 matching score table is finally generated by reshaping the last FC layer. However, the use of a large volume of features in the initial FC layer incurs heavy memory requirements which can lead to a critical computational bottleneck. Hence, we compress the output features from each layer (with a data compression ratio of 16) and feed them to the initial FC layer. We call this process \textit{compression} and the associated components \textit{compressors}.

A \textit{compressor} is composed of three types of layers: 3$\times$3 convolution, ReLU, and Local Response Normalization (LRN). The number of convolution filters is sixteen times smaller than the input channel size. The filter weights of a \textit{compressor} are also shared between the query and search stream. Note that the output value of ReLU at each layer shows a different scale due to the unbounded nonlinearity. These unbalanced scales can cause confusion when the feature instances are globally combined in the FC layers. Moreover, pre-normalized features before a loss function induce more effective network convergence as validated in \Sref{sec:exp_self}. We thus include LRN layer at a \textit{compressor}. Finally, we vectorize and concatenate all of the features from each \textit{compressor} and feed them to the initial FC layer.

The inspiration behind the concept of a \textit{compressor} comes from Wang \etal~\cite{wang2015visual}. They fix the number of channels exploited simply by discarding redundant features from the outputs of the conv4 and conv5 layers. Using such a technique, they proved that a target object can be effectively represented without noise. This can, however, cause a critical loss of valuable information depending on the type of data or the complexity of a scene. Therefore, we adaptively compress our features in order to reduce redundancy without losing reliable information.

\begin{figure*}[t]
	\centering
	\begin{tabular}{@{}c}
		{\includegraphics[width=1\linewidth]{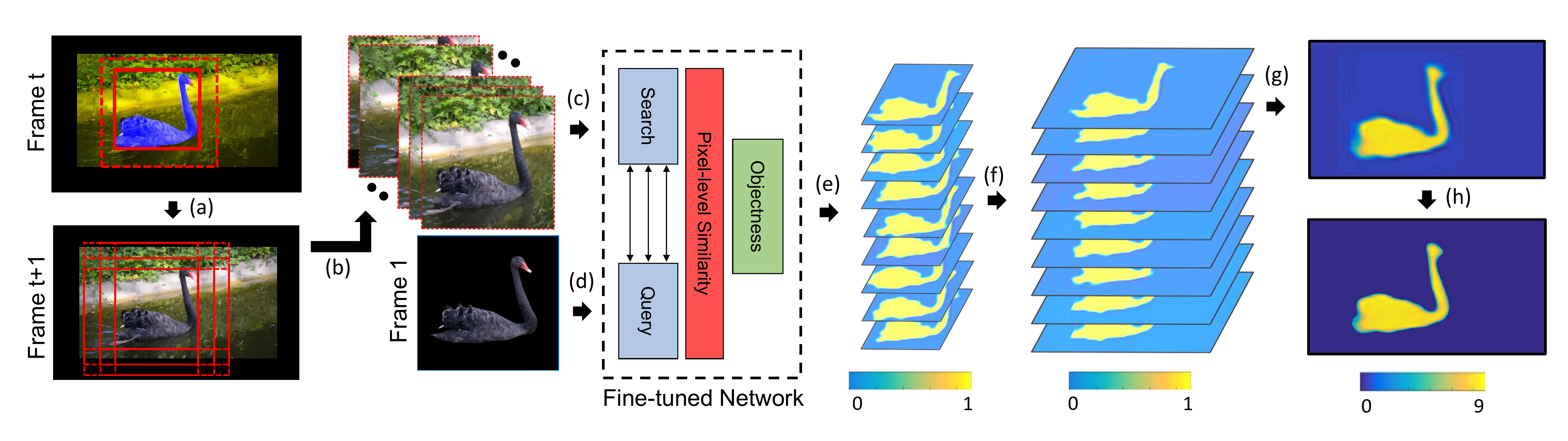}}\hspace{0.1mm}
	\end{tabular}
	\vspace{-3mm}
	\caption{The proposed video object segmentation strategy for an online sequence: (a) Candidates sampling, (b) Image resizing, (c, d) Feeding search and query inputs, (e) Probability maps for each candidate, (f) Size restoration, (g) Stacking, and (h) Thresholding.}
	\label{fig:inference}
	\vspace{-3mm}
\end{figure*}

\textbf{Objectness decoding}\ \ \
The 50$\times$50 matching score table reshaped from the last FC layer encodes the similarity between two input objects. However, direct classification of this matching table at the pixel-level is ineffective due to outliers and inconsistent similarity scores. We thus enforce objectness coherency and reject outliers through several instances of the use of convolutional layers (shaded in green in \Fref{fig:network}). Note that one zero padded input is always convoluted by 3$\times$3 filters to maintain the spatial resolution. An example of a clearly separated target object area is available in \Fref{fig:inference}-(h), and the performance according to the number of decoding layers is validated in \Sref{sec:exp_self}.

\textbf{Loss}\ \ \ 
To constraint the network outputs to binary values (background: 1, target: 0) in each pixel, we use the basic element-wise Euclidean distance as a loss function $\mathfrak{L}$. Let the $P$ indexes be the probability of the output and let $L$ denote the value of the ground-truth label. The score is then estimated by   
\vspace{-2mm}
\begin{equation} 
\label{loss}
\mathfrak{L}=\frac{1}{N^2}\sum_{x=1}^{N}\sum_{y=1}^{N}\parallel P(x,y)-L(x,y) \parallel_{2}, 
\vspace{-2mm}
\end{equation}

where $L\in \{0,\ 1\}$, $N$ is the output size, which is set to 50, and $(x,y)$ is the pixel location in the probability map. It should be noted that Sigmoid activation can be used before L2 loss in order to prevent a "strong positive probability" pixel which results in a large penalty. In our case, however, the normalization layer in each \textit{compressor} effectively pre-adjusts the feature scale to [0, 1], resulting in stable loss convergence with the simple L2 function only (\Fref{fig:ablation}-(b)).

\vspace{-1mm}
\subsection{Video Object Segmentation}\label{sec:inference}
\vspace{-1mm}
\Fref{fig:inference} shows the pipeline of the proposed video object segmentation method for an arbitrary video. Our method to propagate the initial object mask consists of two parts: candidates sampling and optimal area extraction. Before using our pixel-level matching network, however, it must be fine-tuned (second training stage) during the first frame because the network parameters at the end of the pre-training process are optimized exclusively for the objects in the 30 training videos. Thanks to the supervision of the initial target appearance, the model can adapt to any type of object.

\textbf{Initial frame fine-tuning}\ \ \ 
In order to generate the fine-tuning data, we use only the first frame of a new video (not utilized for training). For the query input, we completely segment the target object and crop the bounding box around it with small margins (this step assumes a manually labeled image). For the search input, we sample multiple boxes around the target object with different margins ($\rho_{1}$, $\rho_{2}$, $\rho_{3}$) to deal with various scales. It is further augmented by translation and flipping. Here, the target object is always fully or partially included (at least $50\%$ overlapped) for more efficient weight convergence. The input data (about 150 pairs) are finally resized to 100$\times$100p while the corresponding labels are adjusted to 50$\times$50p. We believe that the feature extraction parts are not highly dependent on the appearance of an arbitrary target. Therefore, the weights for the feature extraction parts (the blue boxes in \Fref{fig:network}), except for the \textit{compressors}, are fixed, whereas the others are updated.

In the field of visual tracking, many previous works \cite{nam2015learning,wang2015visual,wang2015transferring} use a frequent model update strategy in the course of a new video sequence. In our case, however, we do not update the network after fine-tuning for the following reasons. First, we need to consider the computational efficiency for video processing. In consideration of this, the model update is particularly heavy (about ten times slower than one single feed forward process). Secondly, the model update strategy has a rather negative effect on video segmentation because we cannot obtain a perfect pixel-level label in the middle of the sequence. Therefore, updating the model using slightly mislabeled data can increase the drift. Therefore, we match the initial frame with current frame until the end of the sequence without any model update. 

\textbf{Candidates sampling} \ \ \
In order to predict reliable target object area in the following frame, we investigate the network responses to multiple candidates and aggregate their information. To do this, we initially sample nine bounding boxes in the next frame in light of the current target object area with certain margins (\Fref{fig:inference} - (a)). They are then resized into 100$\times$100p patches to fit the input size of the network (\Fref{fig:inference} - (b)). Finally, the patches are fed to the fine-tuned network to obtain the probability maps as shown in \Fref{fig:inference} - (c, d, e).

\textbf{Optimal area extraction}\ \ \ 
Because the size of each probability map is 50$\times$50p, the outputs are reshaped back to the original frame size as described in \Fref{fig:inference}-(f). We then accumulate all of the response maps aligning the pixel locations. Using the combined information, an optimal target area is finally extracted by thresholding with a pre-defined parameter (\Fref{fig:inference}-(h)). This strategy is particularly efficient because the output of the network differs slightly according to the location of the box. Therefore, we can extract a more reliable target object area through the aggregation of multiple pixel responses, especially in the edge area. An investigation of the network responses to a larger number of sample candidates may lead to greater performances. In practice, however, we determined that using nine samples is a good compromise between speed and accuracy.

\vspace{-2mm}
\section{Experiments}\label{sec:exp}
\vspace{-1mm}
In this section, we describe implementation details, and we demonstrate the validity of the proposed network structure. Comparative evaluations on three different benchmarks show the effectiveness of the proposed method. Finally, we apply our method to region tracking tasks using infrared data to demonstrate the transferability of our network to a different domain. 

\vspace{-1mm}
\subsection{Implementation details}\label{sec:exp_param}
\vspace{-1mm}

Our CNN pixel-level matching structure is designed and pre-trained using the \emph{Caffe} toolbox. Our video segmentation algorithm (including fine-tuning) is implemented through the MATLAB wrapper of \emph{Caffe}. All of the experiments are conducted on a desktop computer equipped with a 3.60 GHz CPU and a TITAN X graphic card.

We pre-trained our network using a Stochastic Gradient Descent (SGD) method with a learning rate of $10^{-5}$. The momentum and weight decay parameters are fixed to 0.9 and 5$\times10^{-4}$ respectively. The learning rate is reduced at every 10 epochs, while the parameters of the local size, $\alpha$ and $\beta$ for the LRN are set to 5, $10^{-4}$ and 7.5$\times10^{-1}$ respectively. For fine-tuning using the initial frame, the learning rate is twice as large ($2\times10^{-5}$) for more rapid convergence, while the other parameters are identical to the pre-training parameters. In our experiments, $\rho_1$, $\rho_2$, and $\rho_3$ are fixed at 10, 30 and 50 respectively. After fine-tuning, online model update does not occur during the sequence.

\subsection{Proposed Network Validation}\label{sec:exp_self}
\vspace{-1mm}	
To demonstrate the efficiency of our pixel-level matching network, we conducted a self-structure evaluation on the \textit{DAVIS}~\cite{perazzibenchmark} benchmark with two different metrics: region similarity (intersection-over-union: IoU) and contour accuracy (calculated from the F-measure score). In order to evaluate the system stability, we also report the mean standard deviation of each metric. Here, the speed for one feed forward task in our network is approximately $8\times 10^{-3}$s, while it takes about 1.5$\times 10^{-1}$s per frame to deal with the entire process. The major bottleneck (except for the feed forward time) comes from the image resizing step. In \Tref{tb:davis}, the results from the proposed pixel-level matching network and its post-processed results are denoted as PLM and PLM$_P$, respectively. PLM$_P$ is simply obtained by inserting a mask from PLM with the corresponding image into a publicly available weighted median filter \cite{zhang2014100+} code (filter size 5, iteration 3). It makes the PLM results to be sharper and effectively rejects outliers, while it does not have a significant impact on the computational time.

\begin{figure}[t]			
	\centering
	\begin{tabular}{@{}c}
		{\includegraphics[width=1\linewidth]{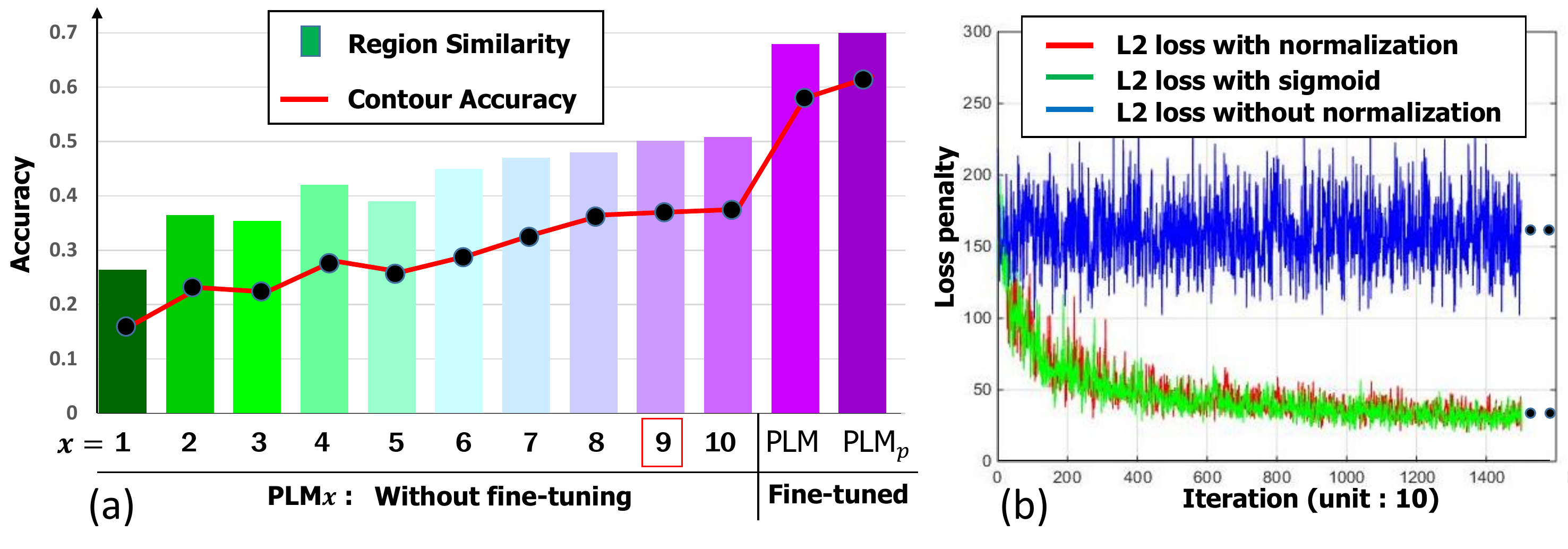}}\hspace{0.1mm}
	\end{tabular}
	
	\caption{(a) The performance graph according to the number of decoding convolutional layers. The structure PLM$_9$ is the identical to the PLM. (b) The loss graphs for three different conditions. }
	\label{fig:ablation}
	\vspace{-3mm}
\end{figure}

\textbf{Ablation test for decoding convolutional (DC) layers}\\ We conduct an ablation test to find the optimal combination of DC layers (the green parts in \Fref{fig:network}). In this assessment, we add DC layers one by one and stop the layer addition process when the accuracy reaches its maximum as depicted in \Fref{fig:ablation}-(a). The performance (in terms of the average score) for each experimental setting (PLM$_1$\~\ PLM$_{10}$) is evaluated on the \textit{DAVIS} 20 test sequences without fine-tuning. Here, the first DC layer increases the channel size from 1 to $2^{(x-1)}$, where $x$ is the total number of DC layers, while the subsequent DC layers reduce the size of the previous channel by a factor of two. From this graph, we can also demonstrate the effect of fine-tuning (second stage training) by comparing PLM$_9$ with PLM because they have identical network structures. 

\textbf{Normalization vs. without normalization}\\
\Fref{fig:ablation}-(b) depicts the effectiveness of the normalization layers in the \textit{compressors}. The loss severely fluctuates without normalization, while the normalized features lead to stable loss convergence even without sigmoid activation. We further noted that this convergence tendency is similar to the case when using "sigmoid activation + L2 loss" function.

\textbf{Use and disuse of \textit{compressor}}\\ 
As described in \Sref{sec:pretraining}, owing to the \textit{compressor}, we could shorten the computational time while also boosting the memory efficiency of the network. In order to highlight the advantages offered by the \textit{compression} technique, we attempted to train the network without \textit{compressors}. However, the memory requirement exceeded 12GB which beyond the capacity of our GPU (even for a batch size of 1). This implies that \textit{compressors} are indispensable when training the network (about 3.7GB with a batch size of 32).

\textbf{Features from multiple layers vs. single layer}\\
To consider the spatial details and semantic information at the same time, we exploit the features from different layers. To underline the importance of this, we compare our network against the same network but using the features from a single layer only denoted as PLM$_S$. In this case, we directly feed the output of the last convolutional layer (in the Siamese structure) to the initial FC layer. We then train and test the network on the same environment. In \Tref{tb:davis}, PLM$_S$ does not work for every sequence due to ineffective pixel-wise object localization, and it is highly affected by cluttered background causing critical drifting effects.

\begin{table}[t]
	\centering
		
	\begin{tabular}[t]{p{1cm}|p{2.2mm}p{2.2mm}p{2.2mm}p{2.2mm}p{2.2mm}p{2.2mm}p{2.2mm}p{4.2mm}|p{3.2mm}p{4mm}}   
		
		\toprule[1.5pt]  
		
		&   \scriptsize{HVS}
		&   \scriptsize{NLC}
		&   \scriptsize{JMP}
		&   \scriptsize{SEA}
		&   \scriptsize{BVS}
     	&   \scriptsize{OFL}
		&   \scriptsize{PLM}
		&  \scriptsize{PLM$_P$} 
		&  \scriptsize{PLM$_S$} 
		&  \scriptsize{PLM$_U$} \\
		
		\midrule[1.5pt]  \\[-3.8ex]
		
		\parbox{0.5cm}{\centering \scriptsize{blackswan} \vspace{-5mm}}
		& \scriptsize{0.91} &\scriptsize{0.87} &\scriptsize{0.93} & \scriptsize{0.93}  &\scriptsize{\textcolor{blue}{\textbf{0.94}}}&\scriptsize{\textcolor{red}{\textbf{0.94}}}& \scriptsize{0.86} &  \scriptsize{0.89} & \scriptsize{0.27}& \scriptsize{0.86}\\[-1ex]
		\parbox{0.8cm}{\centering} 
		&\scriptsize{0.91}&\scriptsize{0.82}&\scriptsize{0.94}&\scriptsize{0.95}&\scriptsize{\textcolor{blue}{\textbf{0.96}}}&\scriptsize{\textcolor{red}{\textbf{0.98}}}&\scriptsize{0.87}&  \scriptsize{0.89} & \scriptsize{0.20}& \scriptsize{0.87}
		\\ [-1ex]\cmidrule{1-11}\\[-3.8ex]
		\parbox{0.5cm}{\centering\scriptsize{bmx} }
		&\scriptsize{0.18} &\scriptsize{0.21} &\scriptsize{0.23} &\scriptsize{0.11} &\scriptsize{0.38}&\scriptsize{0.14} &\scriptsize{\textcolor{blue}{\textbf{0.47}}} &\scriptsize{\textcolor{red}{\textbf{0.48}}} & \scriptsize{0.27}& \scriptsize{0.40}\\[-1ex]
		\parbox{0.5cm}{\centering\scriptsize{tree}} 
		&\scriptsize{0.28}&\scriptsize{0.33}&\scriptsize{0.31}&\scriptsize{0.13}&\scriptsize{0.65}&\scriptsize{0.16}&\scriptsize{\textcolor{blue}{\textbf{0.66}}}&\scriptsize{\textcolor{red}{\textbf{0.68}}} & \scriptsize{0.43}& \scriptsize{0.55}
		\\ [-1ex]\cmidrule{1-11}\\[-3.8ex]
		
		\parbox{0.5cm}{\centering \scriptsize{break}}
		&\scriptsize{\textcolor{blue}{\textbf{0.55}}} &\scriptsize{\textcolor{red}{\textbf{0.67}}} &\scriptsize{0.48}& \scriptsize{0.33}&\scriptsize{0.50}&\scriptsize{0.52}&\scriptsize{0.47}&\scriptsize{0.48}
		& \scriptsize{0.06}& \scriptsize{0.50}	\\[-1ex] 
		\parbox{0.5cm}{\centering\scriptsize{dance}} 
		&\scriptsize{0.47}&\scriptsize{\textcolor{red}{\textbf{0.66}}}&\scriptsize{0.51}&\scriptsize{0.39}&\scriptsize{0.49}&\scriptsize{\textcolor{blue}{\textbf{0.52}}}&\scriptsize{0.35}&\scriptsize{0.41} & \scriptsize{0.08}& \scriptsize{0.36}
		\\[-1ex]  \cmidrule{1-11}\\[-3.8ex]
		\parbox{0.5cm}{\centering \scriptsize{camel}\vspace{-3mm}}   
		&\scriptsize{\textcolor{red}{\textbf{0.87}}} & \scriptsize{0.76}&\scriptsize{0.64}&\scriptsize{0.65}&\scriptsize{0.67}&\scriptsize{\textcolor{blue}{\textbf{0.86}}}&\scriptsize{0.65}&\scriptsize{0.68} & \scriptsize{0.40}& \scriptsize{0.62}     \\[-1ex]
		\parbox{0.5cm}{\centering \scriptsize{}} 
		&\scriptsize{\textcolor{red}{\textbf{0.87}}}&\scriptsize{0.72}&\scriptsize{0.71}&\scriptsize{0.61}&\scriptsize{0.70}&\scriptsize{\textcolor{blue}{\textbf{0.84}}}&\scriptsize{0.54}&\scriptsize{0.61} & \scriptsize{0.20}& \scriptsize{0.49}
		\\[-1ex] \cmidrule{1-11} \\[-3.8ex]    		
		\parbox{0.5cm}{\centering\scriptsize{car}\vspace{-0.7mm}}
		&\scriptsize{0.77}& \scriptsize{0.50}&\scriptsize{0.72}&\scriptsize{0.70}&\scriptsize{0.85}&\scriptsize{\textcolor{red}{\textbf{0.90}}}&\scriptsize{0.86}&\scriptsize{\textcolor{blue}{\textbf{0.87}}}& \scriptsize{0.27}& \scriptsize{0.87}\\[-1ex]
		\parbox{0.5cm}{\centering  \scriptsize{roundabout}\vspace{0.7mm}}   
		&\scriptsize{0.55}&\scriptsize{0.25}&\scriptsize{0.61}&\scriptsize{0.71}&\scriptsize{0.62}&\scriptsize{\textcolor{red}{\textbf{0.76}}}&\scriptsize{0.64}&\scriptsize{\textcolor{blue}{\textbf{0.71}}} & \scriptsize{0.16}& \scriptsize{0.68}\\[-1ex] \cmidrule{1-11}  \\[-3.8ex]		
		\parbox{0.5cm}{\centering \scriptsize{car}\vspace{-0.7mm}}
		&\scriptsize{0.70}&\scriptsize{0.64}& \scriptsize{0.64}&\scriptsize{0.77}&\scriptsize{0.57}&\scriptsize{\textcolor{red}{\textbf{0.84}}}&\scriptsize{0.77}&\scriptsize{\textcolor{blue}{\textbf{0.79}}} & \scriptsize{0.43}& \scriptsize{0.77}\\[-1ex]
		\parbox{0.5cm}{\centering \scriptsize{shadow}}   
		&\scriptsize{0.59}&\scriptsize{0.54}&\scriptsize{0.62}&\scriptsize{\textcolor{red}{\textbf{0.75}}}&\scriptsize{0.47}&\scriptsize{\textcolor{blue}{\textbf{0.74}}}&\scriptsize{0.62}&\scriptsize{0.73}& \scriptsize{0.32}& \scriptsize{0.63}
		\\ [-1ex]\cmidrule{1-11}     \\[-3.8ex]		
		\parbox{0.5cm}{\centering \scriptsize{cows} \vspace{-6mm}}
		&\scriptsize{0.78}&\scriptsize{0.88}&\scriptsize{0.75}&\scriptsize{0.71}&\scriptsize{\textcolor{blue}{\textbf{0.89}}}&\scriptsize{\textcolor{red}{\textbf{0.91}}}&\scriptsize{0.81}&\scriptsize{0.83}& \scriptsize{0.15}& \scriptsize{0.79} \\[-1ex]
		\parbox{0.5cm}{\centering \scriptsize{}}   
		&\scriptsize{0.63}&\scriptsize{0.80}&\scriptsize{0.7}&\scriptsize{0.67}&\scriptsize{\textcolor{red}{\textbf{0.85}}}&\scriptsize{\textcolor{blue}{\textbf{0.85}}}&\scriptsize{0.71}&\scriptsize{0.74}& \scriptsize{0.10}& \scriptsize{0.65}
		\\[-1ex] \cmidrule{1-11}\\[-3.8ex]     		
		\parbox{0.5cm}{\centering \scriptsize{dance}}
		&\scriptsize{0.31}&\scriptsize{0.34}&\scriptsize{0.44}&\scriptsize{0.11}&\scriptsize{0.49}&\scriptsize{0.53}&\scriptsize{\textcolor{blue}{\textbf{0.59}}}&\scriptsize{\textcolor{red}{\textbf{0.62}}}& \scriptsize{0.44}& \scriptsize{0.61}\\[-1ex]
		\parbox{0.5cm}{\centering\scriptsize{twirl}} 
		&\scriptsize{0.51}&\scriptsize{0.36}&\scriptsize{0.52}&\scriptsize{0.21}&\scriptsize{0.48}&\scriptsize{\textcolor{red}{\textbf{0.60}}}&\scriptsize{0.50}&\scriptsize{\textcolor{blue}{\textbf{0.54}}}& \scriptsize{0.35}& \scriptsize{0.49}
		\\[-1ex] \cmidrule{1-11}\\[-3.8ex]  	
		\parbox{0.5cm}{\centering \scriptsize{dog} \vspace{-3mm}}
		&\scriptsize{0.72}&\scriptsize{0.81}&\scriptsize{0.67}&\scriptsize{0.58}&\scriptsize{0.72}&\scriptsize{\textcolor{red}{\textbf{0.89}}}&\scriptsize{0.73}&\scriptsize{\textcolor{blue}{\textbf{0.74}}}& \scriptsize{0.29}& \scriptsize{0.79}\\[-1ex]
		\parbox{0.5cm}{\centering} 
		&\scriptsize{0.63}&\scriptsize{\textcolor{blue}{\textbf{0.70}}}&\scriptsize{0.59}&\scriptsize{0.54}&\scriptsize{0.59}&\scriptsize{\textcolor{red}{\textbf{0.82}}}&\scriptsize{0.61}&\scriptsize{0.63}& \scriptsize{0.15}& \scriptsize{0.63}
		\\[-1ex] \cmidrule{1-11}\\[-3.8ex]  
		\parbox{0.5cm}{\centering \scriptsize{drift}}
		&\scriptsize{0.33}&\scriptsize{0.32}&\scriptsize{0.24}&\scriptsize{0.12}&\scriptsize{0.03}&\scriptsize{0.10}&\scriptsize{\textcolor{red}{\textbf{0.73}}}&\scriptsize{\textcolor{blue}{\textbf{0.71}}}& \scriptsize{0.03}& \scriptsize{0.61}\\[-1ex]
		\parbox{0.5cm}{\centering\scriptsize{chicane}} 
		&\scriptsize{0.54}&\scriptsize{0.31}&\scriptsize{0.33}&\scriptsize{0.16}&\scriptsize{0.07}&\scriptsize{0.21}&\scriptsize{\textcolor{blue}{\textbf{0.79}}}&\scriptsize{\textcolor{red}{\textbf{0.79}}}& \scriptsize{0.03}& \scriptsize{0.60}
		\\[-1ex] \cmidrule{1-11}\\[-3.8ex]  
		\parbox{0.5cm}{\centering \scriptsize{drift}}
		&\scriptsize{0.29}&\scriptsize{0.47}&\scriptsize{0.61}&\scriptsize{0.51}&\scriptsize{0.40}&\scriptsize{0.33}&\scriptsize{\textcolor{blue}{\textbf{0.73}}}&\scriptsize{\textcolor{red}{\textbf{0.74}}}& \scriptsize{0.12}& \scriptsize{0.63}\\[-1ex]
		\parbox{0.5cm}{\centering\scriptsize{straight}}
		&\scriptsize{0.26}&\scriptsize{0.38}&\scriptsize{0.47}&\scriptsize{0.50}&\scriptsize{0.41}&\scriptsize{0.27}&\scriptsize{\textcolor{blue}{\textbf{0.52}}}&\scriptsize{\textcolor{red}{\textbf{0.56}}}& \scriptsize{0.05}& \scriptsize{0.50}
		\\[-1ex] \cmidrule{1-11}\\[-3.8ex]  
		\parbox{0.5cm}{\centering \scriptsize{goat}\vspace{-3mm}}
		&\scriptsize{0.58}&\scriptsize{0.01}&\scriptsize{0.73}&\scriptsize{0.53}&\scriptsize{0.66}&\scriptsize{\textcolor{red}{\textbf{0.86}}}&\scriptsize{0.76}&\scriptsize{\textcolor{blue}{\textbf{0.77}}}& \scriptsize{0.05}& \scriptsize{0.74}
		\\[-1ex]
		\parbox{0.5cm}{\centering} 
		&\scriptsize{0.54}&\scriptsize{0.13}&\scriptsize{0.61}&\scriptsize{0.47}&\scriptsize{0.58}&\scriptsize{\textcolor{red}{\textbf{0.84}}}&\scriptsize{0.63}&\scriptsize{\textcolor{blue}{\textbf{0.69}}}& \scriptsize{0.05}& \scriptsize{0.60}
		\\[-1ex] \cmidrule{1-11}\\[-3.8ex]  
		\parbox{0.5cm}{\centering \scriptsize{horse}}
		&\scriptsize{0.76}&\scriptsize{\textcolor{blue}{\textbf{0.83}}}&\scriptsize{0.58}&\scriptsize{0.63}&\scriptsize{0.80}&\scriptsize{\textcolor{red}{\textbf{0.86}}}&\scriptsize{0.72}&\scriptsize{0.77}& \scriptsize{0.38}& \scriptsize{0.71}\\[-1ex]
		\parbox{0.5cm}{\centering\scriptsize{jump}\vspace{0.3mm}} 
		&\scriptsize{0.80}&\scriptsize{\textcolor{blue}{\textbf{0.88}}}&\scriptsize{0.65}&\scriptsize{0.65}&\scriptsize{0.80}&\scriptsize{\textcolor{red}{\textbf{0.90}}}&\scriptsize{0.73}&\scriptsize{0.78}& \scriptsize{0.24}& \scriptsize{0.70}
		\\[-1ex] \cmidrule{1-11}\\[-3.8ex]  
		\parbox{0.5cm}{\centering \scriptsize{kite}\vspace{-0.7mm}}
		&\scriptsize{0.40}&\scriptsize{0.45}&\scriptsize{0.50}&\scriptsize{0.48}&\scriptsize{0.42}&\scriptsize{\textcolor{red}{\textbf{0.70}}}&\scriptsize{0.59}&\scriptsize{\textcolor{blue}{\textbf{0.64}}}& \scriptsize{0.20}& \scriptsize{0.25}\\[-1ex]
		\parbox{0.5cm}{\centering\scriptsize{surf}} 
		&\scriptsize{0.37}&\scriptsize{0.45}&\scriptsize{0.31}&\scriptsize{0.28}&\scriptsize{\textcolor{blue}{\textbf{0.64}}}&\scriptsize{\textcolor{blue}{\textbf{0.49}}}&\scriptsize{0.45}&\scriptsize{0.45}& \scriptsize{0.15}& \scriptsize{0.17}
		\\[-1ex] \cmidrule{1-11}\\[-3.8ex]  
		\parbox{0.5cm}{\centering \scriptsize{libby}\vspace{-3mm}}
		&\scriptsize{0.55}&\scriptsize{\textcolor{blue}{\textbf{0.63}}}&\scriptsize{0.29}&\scriptsize{0.22}&\scriptsize{\textcolor{red}{\textbf{0.77}}}&\scriptsize{0.55}&\scriptsize{0.52}&\scriptsize{0.54}& \scriptsize{0.23}& \scriptsize{0.43}\\[-1ex]
		\parbox{0.5cm}{\centering} 
		&\scriptsize{0.64}&\scriptsize{\textcolor{red}{\textbf{0.74}}}&\scriptsize{0.36}&\scriptsize{0.21}&\scriptsize{\textcolor{blue}{\textbf{0.84}}}&\scriptsize{0.61}&\scriptsize{0.57}&\scriptsize{0.58}& \scriptsize{0.21}& \scriptsize{0.45}
		\\[-1ex] \cmidrule{1-11}\\[-3.8ex]  
		\parbox{0.5cm}{\centering \scriptsize{motorcross}\vspace{-0.7mm}}
		&\scriptsize{0.09}&\scriptsize{0.25}&\scriptsize{\textcolor{blue}{\textbf{0.58}}}&\scriptsize{0.38}&\scriptsize{0.34}&\scriptsize{\textcolor{red}{\textbf{0.60}}}&\scriptsize{0.49}&\scriptsize{0.51}& \scriptsize{0.27}& \scriptsize{0.44}\\[-1ex]
		\parbox{0.5cm}{\centering\scriptsize{jump} } 
		&\scriptsize{0.13}&\scriptsize{0.30}&\scriptsize{\textcolor{red}{\textbf{0.54}}}&\scriptsize{0.40}&\scriptsize{0.37}&\scriptsize{\textcolor{blue}{\textbf{0.47}}}&\scriptsize{0.32}&\scriptsize{0.37}& \scriptsize{0.06}& \scriptsize{0.27}
		\\[-1ex]
		\cmidrule{1-11}\\[-3.8ex]  
		\parbox{0.5cm}{\centering \scriptsize{paragliding}\vspace{-0.7mm}}
		&\scriptsize{0.53}&\scriptsize{0.62}&\scriptsize{0.59}&\scriptsize{0.57}&\scriptsize{\textcolor{red}{\textbf{0.64}}}&\scriptsize{\textcolor{blue}{\textbf{0.63}}}&\scriptsize{0.55}&\scriptsize{0.57}& \scriptsize{0.41}& \scriptsize{0.54}\\[-1ex]
		\parbox{0.5cm}{\centering\scriptsize{launch}} 
		&\scriptsize{0.20}&\scriptsize{0.24}&\scriptsize{0.17}&\scriptsize{0.18}&\scriptsize{\textcolor{red}{\textbf{0.32}}}&\scriptsize{\textcolor{blue}{\textbf{0.25}}}&\scriptsize{0.15}&\scriptsize{0.18}& \scriptsize{0.11}& \scriptsize{0.16}
		\\[-1ex] \cmidrule{1-11}\\[-3.8ex]  
		\parbox{0.5cm}{\centering \scriptsize{parkour}\vspace{-3mm}}
		&\scriptsize{0.24}&\scriptsize{\textcolor{red}{\textbf{0.90}}}&\scriptsize{0.34}&\scriptsize{0.12}&\scriptsize{0.75}&\scriptsize{\textcolor{blue}{\textbf{0.85}}}&\scriptsize{0.77}&\scriptsize{0.82}& \scriptsize{0.05}& \scriptsize{0.57}\\[-1ex]
		\parbox{0.5cm}{\centering} 
		&\scriptsize{0.32}&\scriptsize{\textcolor{red}{\textbf{0.91}}}&\scriptsize{0.41}&\scriptsize{0.27}&\scriptsize{0.67}&\scriptsize{\textcolor{blue}{\textbf{0.87}}}&\scriptsize{0.75}&\scriptsize{0.81}& \scriptsize{0.08}& \scriptsize{0.56}
		\\[-1ex] \cmidrule{1-11}\\[-3.8ex]  
		\parbox{0.5cm}{\centering \scriptsize{scooter}}
		&\scriptsize{0.62}&\scriptsize{0.16}&\scriptsize{0.62}&\scriptsize{\textcolor{blue}{\textbf{0.79}}}&\scriptsize{0.33}&\scriptsize{\textcolor{red}{\textbf{0.80}}}&\scriptsize{0.73}&\scriptsize{0.75}& \scriptsize{0.09}& \scriptsize{0.69}\\[-1ex]
		\parbox{0.5cm}{\centering\scriptsize{black}} 
		&\scriptsize{0.57}&\scriptsize{0.22}&\scriptsize{0.53}&\scriptsize{\textcolor{blue}{\textbf{0.72}}}&\scriptsize{0.40}&\scriptsize{\textcolor{red}{\textbf{0.73}}}&\scriptsize{0.57}&\scriptsize{0.64}& \scriptsize{0.10}& \scriptsize{0.53}
		\\[-1ex] \cmidrule{1-11}\\[-3.8ex]  
		\parbox{0.5cm}{\centering \scriptsize{soapbox}\vspace{-3mm}}
		&\scriptsize{0.68}&\scriptsize{0.63}&\scriptsize{0.75}&\scriptsize{\textcolor{blue}{\textbf{0.78}}}&\scriptsize{\textcolor{red}{\textbf{0.79}}}&\scriptsize{0.68}&\scriptsize{0.72}&\scriptsize{0.76}& \scriptsize{0.16}& \scriptsize{0.70}\\[-1ex]
		\parbox{0.5cm}{\centering} 
		&\scriptsize{0.69}&\scriptsize{0.65}&\scriptsize{0.67}&\scriptsize{\textcolor{blue}{\textbf{0.75}}}&\scriptsize{\textcolor{red}{\textbf{0.76}}}&\scriptsize{0.67}&\scriptsize{0.54}&\scriptsize{0.62}& \scriptsize{0.18}& \scriptsize{0.49}
		\\[-1ex] \cmidrule{1-11}\\[-3.8ex]  
		\midrule[1.5pt]\\[-3.8ex]  			
		\parbox{0.5cm}{\centering \scriptsize{Average}\vspace{-3mm}}
		&\scriptsize{0.54}&\scriptsize{0.55}&\scriptsize{0.56}&\scriptsize{0.50}&\scriptsize{0.59}&\scriptsize{0.67}&\scriptsize{\textcolor{blue}{\textbf{0.68}}}&\scriptsize{\textcolor{red}{\textbf{0.70}}} & \scriptsize{0.23}& \scriptsize{0.63}\\[-1ex]
		\parbox{1.2cm}{\centering}
		&\scriptsize{0.52}&\scriptsize{0.52}&\scriptsize{0.53}&\scriptsize{0.47}&\scriptsize{0.58}&\scriptsize{\textcolor{red}{\textbf{0.63}}}&\scriptsize{0.58}&\scriptsize{\textcolor{blue}{\textbf{0.62}}}& \scriptsize{0.16}& \scriptsize{0.52}
		\\[-1ex]
		\midrule[1.2pt]\\[-3.8ex] 
		\parbox{0.5cm}{\centering
		\scriptsize{Stability}\vspace{-3mm}}
	&\scriptsize{0.24}&\scriptsize{0.26}&\scriptsize{0.18}&\scriptsize{0.25}&\scriptsize{0.13}&\scriptsize{0.25}&\scriptsize{\textcolor{blue}{\textbf{0.12}}}&\scriptsize{\textcolor{red}{\textbf{0.12}}}&\scriptsize{0.13} & \scriptsize{0.16}\\[-1ex]
	\parbox{1.2cm}{\centering}
	&\scriptsize{0.21}&\scriptsize{0.24}&\scriptsize{0.18}&\scriptsize{0.24}&\scriptsize{0.21}&\scriptsize{0.25}&\scriptsize{\textcolor{blue}{\textbf{0.17}}}&\scriptsize{\textcolor{red}{\textbf{0.16}}}&\scriptsize{0.10} & \scriptsize{0.17}\\[-1ex]

		\midrule[1.2pt]\\[-2.8ex] 
		\parbox{0.5cm}{\centering \scriptsize{Speed}}
		&\scriptsize{5s}&\scriptsize{20s}&\scriptsize{12s}&\scriptsize{6s}&\scriptsize{0.37s}&\scriptsize{42.2s}&\scriptsize{\textcolor{red}{\textbf{0.15s}}}&\scriptsize{\textcolor{blue}{\textbf{0.3s}}}& \scriptsize{0.2s}& \scriptsize{3.85s} \\[0ex]

		\bottomrule[2pt]  
	\end{tabular}     
	\vspace{0mm}
	\caption{Performance evaluation on the \textit{DAVIS} \cite{perazzibenchmark} benchmark. First: IoU score for region similarity (higher is better). Second: F-measure (higher is better) for contour accuracy. Stability is denoted by the mean standard deviation of each metric (lower is better). At the end of the table, we present the approximate processing time for each method as reported in earlier work~\cite{marki2016bilateral}. Here, the left part is for a comparative evaluation, while the right one is for the self-structure evaluation. \color{red}{\textbf{Red}}\color{black}{: best}, \color{blue}{\textbf{blue}}\color{black}{: second best.} For comparison, the following baseline algorithms are assessed: SEA: SeamSeg \cite{ramakanth2014seamseg}, JMP: JumpCut \cite{fan2015jumpcut}, NLC: Non-Local Consensus Voting \cite{faktor2014video}, HVS: Efficient Hierarchical Graph-Based Video Segmentation \cite{grundmann2010efficient}, BVS: Bilateral Space Video Segmentation \cite{marki2016bilateral}, OFL: Video Segmentation via Object Flow \cite{Tsai_CVPR_2016}.} 
	\label{tb:davis}
	\vspace{-4.2mm}
\end{table}

\textbf{Initial frame matching without a model update vs. successive frames matching with a model updates}\\
To prevent the background drift problem and ensure good computational efficiency, our strategy consists of matching the initially fine-tuned query input with the entire sequence. In order to validate the effectiveness of our strategy, we compare it against successive frame matching with a model update scheme. To do so, the query input for frame $t+1$ is generated using the output mask of the previous frame $t$. The network model is updated every ten frames through 50 iterations. Here, the model update is done in a manner similar to that in our fine-tuning scheme (described in \Sref{sec:inference}). The performance of the pixel-level matching model with updating is denoted as PLM$_U$. In \Tref{tb:davis}, PLM$_U$ is comparable to PLM, but it never outperforms PLM. This is mainly due to the fact that model updating and matching between two successive frames using continuously drifted outputs lead to rather degenerated results.

\vspace{-1mm}	
\subsection{Comparative Evaluation}\label{comparison}
\vspace{-1mm}	
 In our comparative experiments, we measure how much the key frame object mask is precisely propagated through the sequences on four different benchmarks.  

\begin{figure}[t]
	\centering
	\begin{tabular}{@{}c}
		{\includegraphics[width=1\linewidth]{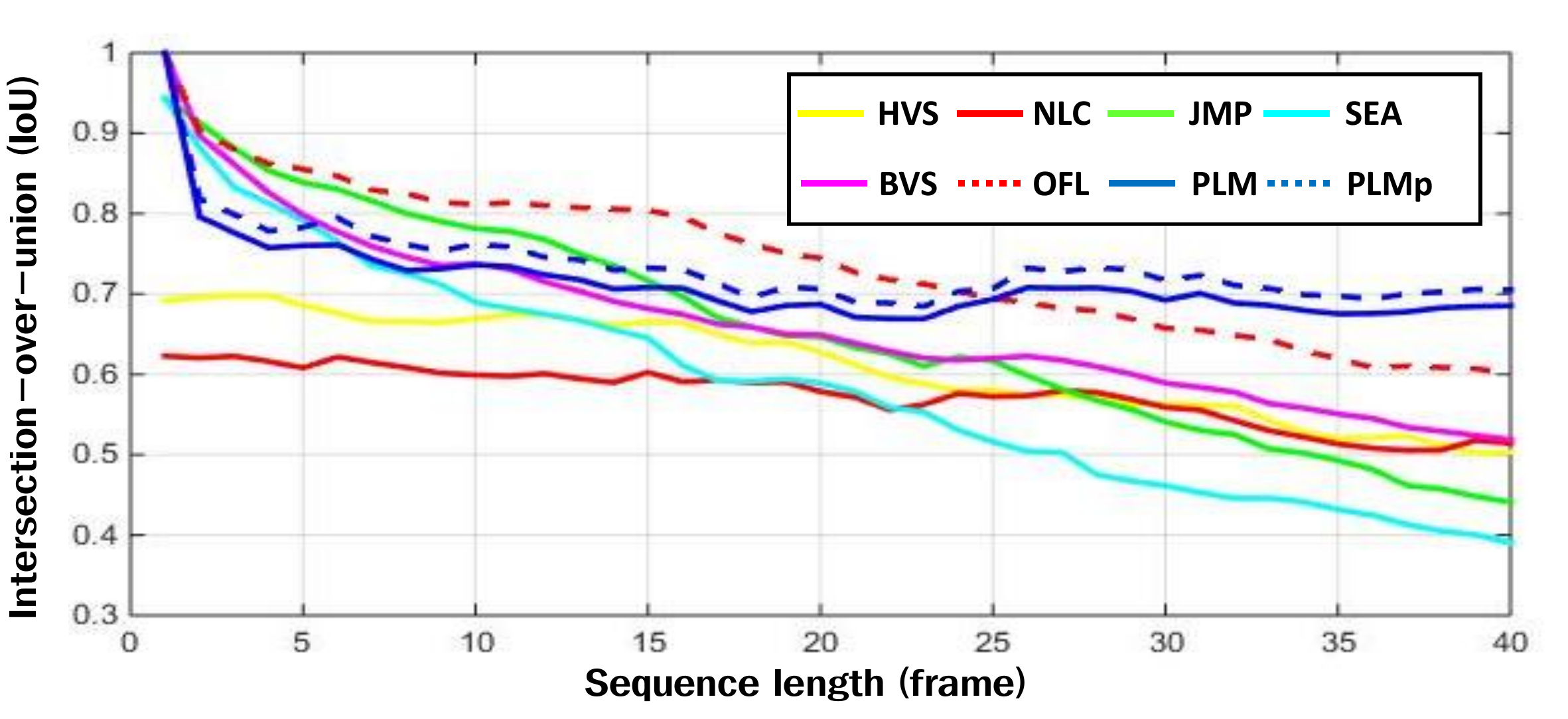}}\hspace{0.1mm}
	\end{tabular}
	
	\caption{The average IoU graph at each sequence as evaluated on the \textit{DAVIS}~\cite{perazzibenchmark} benchmark. The performances of most baselines decrease except for NLC as the initial object mask is propagated. Due to the low dependency between successive frames, our method shows stable graph appearance with high accuracy despite the fact that the initial mask reaches the back part of the sequence. }
	\label{fig:graph}
	\vspace{-3mm}
\end{figure}
\begin{table}[b]
		\vspace{-3mm}			
		
	\centering
	\begin{tabular}[t]{p{1cm}|p{3mm}p{3mm}p{3mm}p{3mm}p{3mm}p{3mm}p{3mm}p{3mm}p{5mm}}   
		
		\toprule[1.5pt]  
		
		&   \scriptsize{HSV}
		&   \scriptsize{FST}
		&   \scriptsize{DAG}
		&   \scriptsize{TMF}
		&   \scriptsize{KEY}
		&   \scriptsize{NLC}
		&  \scriptsize{BVS}
		&  \scriptsize{PLM}
		&  \scriptsize{PLM$_P$}\\
		\midrule[1.5pt]  \\[-3.2ex]
		
		
		\parbox{0.5cm}{\centering \scriptsize{birdfall}}
		&\scriptsize{0.57}&\scriptsize{0.59}&\scriptsize{\textcolor{blue}{\textbf{0.71}}}&\scriptsize{0.62}&\scriptsize{0.49}&\scriptsize{\textcolor{red}{\textbf{0.74}}}&  \scriptsize{0.66}&  \scriptsize{0.64}&  \scriptsize{0.65}
		\\ [-0.5ex]\cmidrule{1-10}\\[-3.2ex]
		\parbox{0.5cm}{\centering\scriptsize{cheetah}} 
		&\scriptsize{0.19}&\scriptsize{0.28}&\scriptsize{0.40}&\scriptsize{0.37}&\scriptsize{0.44}&\scriptsize{\textcolor{red}{\textbf{0.69}}}	&  \scriptsize{0.10}&  \scriptsize{0.65}&  \scriptsize{\textcolor{blue}{\textbf{0.65}}}
		\\ [-0.5ex]\cmidrule{1-10}\\[-3.2ex]
		
		\parbox{0.5cm}{\centering \scriptsize{girl}}
		&\scriptsize{0.32}&\scriptsize{0.73}&\scriptsize{0.82}&\scriptsize{0.89}&\scriptsize{0.88}&\scriptsize{\textcolor{red}{\textbf{0.91}}}&  \scriptsize{\textcolor{blue}{\textbf{0.89}}}&  \scriptsize{0.77}&  \scriptsize{0.78}
		\\[-0.5ex]  \cmidrule{1-10}\\[-3.2ex]
		\parbox{0.5cm}{\centering \scriptsize{monkeydog}}   
		&\scriptsize{0.68}&\scriptsize{\textcolor{red}{\textbf{0.79}}}&\scriptsize{\textcolor{blue}{\textbf{0.75}}}&\scriptsize{0.71}&\scriptsize{0.74}&\scriptsize{0.78}	&  \scriptsize{0.41}&  \scriptsize{0.61}&  \scriptsize{0.72} 
		\\[-0.5ex] \cmidrule{1-10} \\[-3.2ex]    		 	
		\parbox{0.5cm}{\centering\scriptsize{parachute}} 
		&\scriptsize{0.69}&\scriptsize{0.91}&\scriptsize{0.94}&\scriptsize{0.93}&\scriptsize{\textcolor{red}{\textbf{0.96}}}&\scriptsize{0.94}&  \scriptsize{\textcolor{blue}{\textbf{0.94}}}&  \scriptsize{0.85}&  \scriptsize{0.88}
		\\[-0.5ex] \midrule[1.5pt]\\[-3.2ex]  	
		\parbox{0.5cm}{\centering \scriptsize{Average}}
		&\scriptsize{0.49}&\scriptsize{0.66}&\scriptsize{0.72}&\scriptsize{0.70}&\scriptsize{0.70}&\scriptsize{\textcolor{red}{\textbf{0.81}}}&  \scriptsize{0.60}&  \scriptsize{0.70}&  \scriptsize{\textcolor{blue}{\textbf{0.73}}}
		\\[-0.5ex]
		\bottomrule[2pt]  
	\end{tabular}   
	\vspace{1.5mm} 
	\caption{Performance evaluation on the \textit{SegTrack} \cite{FliICCV2013} benchmark using IoU score (higher is better). \color{red}{\textbf{Red}}\color{black}{: best} , \color{blue}{\textbf{blue}}\color{black}{: second best.}}
	
	\label{tb:segtrack}
	
\end{table}	

\textbf{\textit{DAVIS}}~\cite{perazzibenchmark} supplies 30 training datasets and 20 test datasets. This database provides a wide range of challenging scenarios, such as occlusions, dynamic deformations, illumination changes, and scale variations. Note that we did not use any additional datasets to train the network even when testing on different benchmarks.

\begin{table}[t]
	\centering
	
	\begin{tabular}[t]{p{0.3cm}p{2.5mm}|p{7mm}p{7mm}p{7mm}p{7mm}p{7mm}p{7mm}}   
		
		\toprule[1.5pt]\\[-3.5ex]  		
		&{}	
		&  \scriptsize{RB} 
		&   \scriptsize{DA}
		&   \scriptsize{SEA}
		&   \scriptsize{JMP}
		&   \scriptsize{PLM}
		&   {\scriptsize{PLM$_{p}$}}
		
		\\
		\\[-6ex]	\midrule[1.5pt]  \\[-3.8ex]
		
		
		\parbox{0.5cm}{\centering \scriptsize{animation}}
		&{}
		& \scriptsize{11.9} &\scriptsize{6.38} &\scriptsize{6.78} & \scriptsize{\textcolor{red}{\textbf{4.55}}} &\scriptsize{9.38}& \scriptsize{\textcolor{blue}{\textbf{5.86}}}\\[-1ex]	\cmidrule{1-8}\\[-3.5ex]

		\parbox{0.5cm}{\centering\scriptsize{bball} }
		&{}
		& \scriptsize{18.4} &\scriptsize{8.47} &\scriptsize{8.89} & \scriptsize{\textcolor{red}{\textbf{3.90}}} &\scriptsize{12.8}& \scriptsize{\textcolor{blue}{\textbf{8.04}}}
		\\[-1ex]
		\cmidrule{1-8}\\[-3.5ex]
		
		\parbox{0.5cm}{\centering \scriptsize{bear}}
		&{}
		& \scriptsize{4.58} &\scriptsize{4.48} &\scriptsize{4.21} & \scriptsize{\textcolor{blue}{\textbf{4.00}}} &\scriptsize{8.60}& \scriptsize{\textcolor{red}{\textbf{3.45}}}
		\\[-1ex]
		\cmidrule{1-8}\\[-3.5ex]
		\parbox{0.5cm}{\centering \scriptsize{car}}   
		&{}	
		
		& \scriptsize{\textcolor{red}{\textbf{1.76}}} &\scriptsize{5.93} &\scriptsize{5.08} & \scriptsize{2.26} &\scriptsize{4.12}& \scriptsize{\textcolor{blue}{\textbf{2.18}}}
		\\[-1ex]
		\cmidrule{1-8}\\[-3.5ex]	
		\parbox{0.5cm}{\centering\scriptsize{cheetah}}
		&{}	
		& \scriptsize{31.5} &\scriptsize{16.6} &\scriptsize{\textcolor{red}{\textbf{7.68}}} & \scriptsize{\textcolor{blue}{\textbf{8.16}}} &\scriptsize{14.1}& \scriptsize{11.8}
		
		\\[-2.8ex]
		\cmidrule{1-8}\\[-3.5ex]	
		\parbox{0.5cm}{\centering \scriptsize{couple}}
		&{}
		
		& \scriptsize{17.5} &\scriptsize{16.0} &\scriptsize{23.4} & \scriptsize{\textcolor{red}{\textbf{5.13}}} &\scriptsize{13.1}& \scriptsize{\textcolor{blue}{\textbf{9.14}}}
		\\[-1ex]
		\cmidrule{1-8}\\[-3.5ex]	
		
		\parbox{0.5cm}{\centering \scriptsize{cup}}
		&{}	
		& \scriptsize{\textcolor{blue}{\textbf{5.45}}} &\scriptsize{12.9} &\scriptsize{9.31} & \scriptsize{\textcolor{red}{\textbf{2.15}}} &\scriptsize{8.63}& \scriptsize{6.04}
		\\[-1ex]
		\cmidrule{1-8}\\[-3.5ex]

		\parbox{0.5cm}{\centering \scriptsize{dance}}
		&{}
		& \scriptsize{56.1} &\scriptsize{50.8} &\scriptsize{43.0} & \scriptsize{\textcolor{blue}{\textbf{18.7}}} &\scriptsize{31.5}& \scriptsize{\textcolor{red}{\textbf{14.7}}}
		\\[-1ex]
		\cmidrule{1-8}\\[-3.5ex]
		
		\parbox{0.5cm}{\centering \scriptsize{fish}}
		&{}	
		& \scriptsize{51.8} &\scriptsize{21.7} &\scriptsize{25.7} & \scriptsize{17.5} &\scriptsize{\textcolor{blue}{\textbf{9.39}}}& \scriptsize{\textcolor{red}{\textbf{7.42}}}
		\\[-1ex]
		\cmidrule{1-8}\\[-3.5ex]
		\parbox{0.5cm}{\centering \scriptsize{giraffe}}
		&{}
		& \scriptsize{22.0} &\scriptsize{\textcolor{blue}{\textbf{11.2}}} &\scriptsize{17.4} & \scriptsize{\textcolor{red}{\textbf{7.40}}} &\scriptsize{19.7}& \scriptsize{17.4}
		\\[-1ex]
		\cmidrule{1-8}\\[-3.5ex]
		\parbox{0.5cm}{\centering \scriptsize{goat}}
		&{}
		& \scriptsize{13.1} &\scriptsize{13.3} &\scriptsize{\textcolor{blue}{\textbf{8.22}}} & \scriptsize{\textcolor{red}{\textbf{4.14}}} &\scriptsize{17.8}& \scriptsize{15.2}
		\\[-1ex]
		\cmidrule{1-8}\\[-3.5ex]
		\parbox{0.5cm}{\centering \scriptsize{hiphop}}
		&{}
		& \scriptsize{67.5} &\scriptsize{51.1} &\scriptsize{33.7} & \scriptsize{\textcolor{blue}{\textbf{14.2}}} &\scriptsize{19.9}& \scriptsize{\textcolor{red}{\textbf{13.6}}}
		\\[-1ex]
		\cmidrule{1-8}\\[-3.5ex]
		\parbox{0.5cm}{\centering \scriptsize{horse}}
		&{}	
		& \scriptsize{8.39} &\scriptsize{45.1} &\scriptsize{37.8} & \scriptsize{\textcolor{red}{\textbf{6.80}}} &\scriptsize{11.5}& \scriptsize{\textcolor{blue}{\textbf{7.94}}}
		\\[-1ex]
		\cmidrule{1-8}\\[-3.5ex]
		\parbox{0.5cm}{\centering \scriptsize{kongfu}}
		&{}
		& \scriptsize{40.2} &\scriptsize{40.8} &\scriptsize{17.9} & \scriptsize{\textcolor{blue}{\textbf{8.00}}} &\scriptsize{9.74}& \scriptsize{\textcolor{red}{\textbf{6.25}}}
		\\[-1ex]
		\cmidrule{1-8}\\[-3.5ex]		
		\parbox{0.5cm}{\centering \scriptsize{park}}
		&{}
		& \scriptsize{11.8} &\scriptsize{\textcolor{blue}{\textbf{6.54}}} &\scriptsize{6.91} & \scriptsize{\textcolor{red}{\textbf{5.39}}} &\scriptsize{16.5}& \scriptsize{10.2}
		\\[-1ex]
		\cmidrule{1-8}\\[-3.5ex]
		\parbox{0.5cm}{\centering \scriptsize{pig}}
		&{}	
		& \scriptsize{9.22} &\scriptsize{9.85} &\scriptsize{10.3} & \scriptsize{\textcolor{red}{\textbf{3.43}}} &\scriptsize{9.09}& \scriptsize{\textcolor{blue}{\textbf{5.15}}}
		\\[-1ex]
		\cmidrule{1-8}\\[-3.5ex]
		\parbox{0.5cm}{\centering \scriptsize{pot}}
		&{}
		& \scriptsize{\textcolor{red}{\textbf{2.43}}} &\scriptsize{5.03} &\scriptsize{2.98} & \scriptsize{2.95} &\scriptsize{5.46}& \scriptsize{\textcolor{blue}{\textbf{2.66}}}
		\\[-1ex]
		\cmidrule{1-8}\\[-3.5ex]
		\parbox{0.5cm}{\centering \scriptsize{skater}}
		&{}
		& \scriptsize{38.7} &\scriptsize{40.8} &\scriptsize{29.6} & \scriptsize{22.8} &\scriptsize{\textcolor{blue}{\textbf{16.8}}}& \scriptsize{\textcolor{red}{\textbf{12.6}}}
		\\[-1ex]
		\cmidrule{1-8}\\[-3.5ex]
		\parbox{0.5cm}{\centering \scriptsize{station}}
		&{}
		& \scriptsize{8.85} &\scriptsize{20.9} &\scriptsize{21.3} & \scriptsize{9.01} &\scriptsize{\textcolor{blue}{\textbf{7.31}}}& \scriptsize{\textcolor{red}{\textbf{4.68}}}
		\\[-1ex]
		\cmidrule{1-8}\\[-3.5ex]
		\parbox{0.5cm}{\centering \scriptsize{supertramp}}
		&{}	
		& \scriptsize{129} &\scriptsize{60.5} &\scriptsize{57.4} & \scriptsize{42.9} &\scriptsize{\textcolor{blue}{\textbf{30.4}}}& \scriptsize{\textcolor{red}{\textbf{20.7}}}
		\\[-1ex]
		\cmidrule{1-8}\\[-3.5ex]
		\parbox{0.5cm}{\centering \scriptsize{toy}}
		&{}
		& \scriptsize{\textcolor{red}{\textbf{1.28}}} &\scriptsize{3.19} &\scriptsize{2.16} & \scriptsize{\textcolor{blue}{\textbf{1.30}}} &\scriptsize{5.07}& \scriptsize{2.25}
		\\[-1ex]
		\cmidrule{1-8}\\[-3.5ex]
		\parbox{0.5cm}{\centering \scriptsize{tricking}}
		&{}
		& \scriptsize{79.4} &\scriptsize{70.9} &\scriptsize{35.8} & \scriptsize{\textcolor{blue}{\textbf{21.3}}} &\scriptsize{21.9}& \scriptsize{\textcolor{red}{\textbf{15.7}}}
		\\[-1ex]
		
		\midrule[1.5pt]\\[-3.8ex]  			
		\parbox{0.5cm}{\centering \scriptsize{Average} \vspace{-3mm}}
		&{}
		& \scriptsize{28.6} &\scriptsize{23.7} &\scriptsize{18.8} & \scriptsize{\textcolor{blue}{\textbf{9.81}}} &\scriptsize{13.7}& \scriptsize{\textcolor{red}{\textbf{9.55}}} \\[-0.5ex]

		\bottomrule[2pt]  
	\end{tabular}    
	\vspace{1mm}
	\caption{Errors on the \textit{JumpCut} \cite{fan2015jumpcut} benchmark using a transfer distance of sixteen (lower is better). \color{red}{\textbf{Red}}\color{black}{: best} , \color{blue}{\textbf{blue}}\color{black}{: second best.}}
	
	\label{tb:jumpcut}
	\vspace{-4mm}
\end{table}

\begin{figure*}[t]
	\centering
	\begin{tabular}{@{}c}
		{\includegraphics[width=1\linewidth]{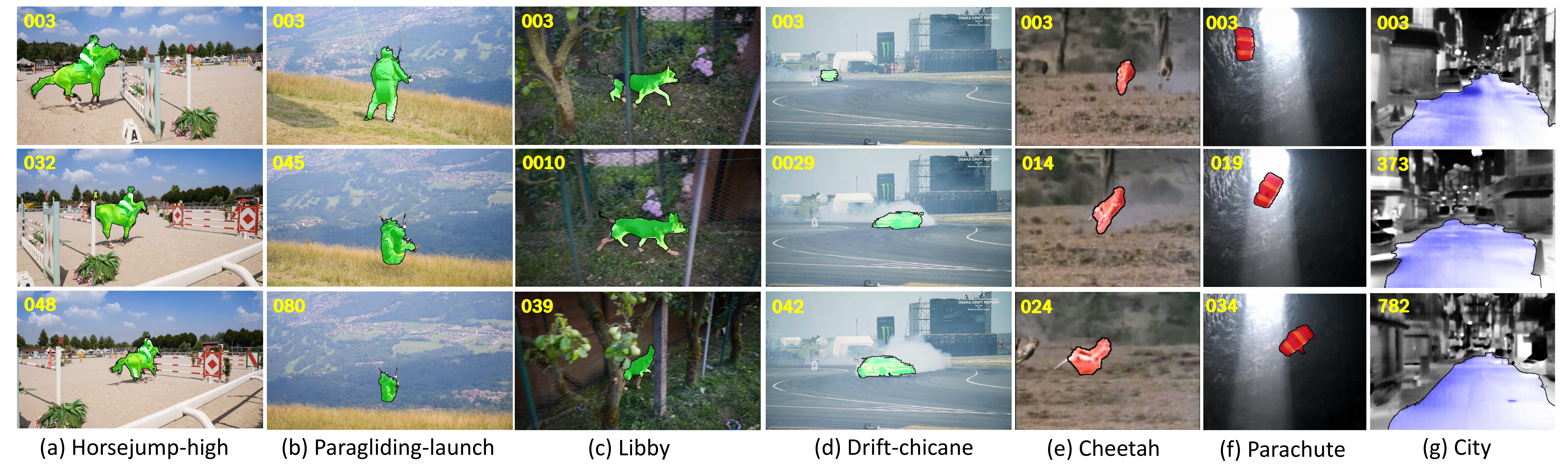}}\hspace{0.1mm}
	\end{tabular}
	
	\caption{Representative results of the proposed method on challenging scenarios. Each color of the output mask is associated with each benchmark: {\color{green}{green}}, {\color{red}{red}} and {\color{blue}{blue}} come from \textit{DAVIS} \cite{perazzibenchmark}, \textit{SegTrack} \cite{FliICCV2013} and \textit{Thermal-Road} \cite{yoon2016thermal} benchmark respectively.}
	\label{fig:quality}
	\vspace{-2mm}
\end{figure*}

As described in \Tref{tb:davis}, the proposed method generally shows better performance on average compared to all of the other approaches. Among the baselines, OFL is highly comparable but our method is much more efficient in terms of the processing time and system stability which are crucial components in real-world situations. Overall, there are two reasons for these promising results. First, our pixel-level matching network can encompass semantic-level information as well as spatial details. Therefore, our method robustly responds to many challenging scenarios such as occlusions, illumination changes, scale variations, and non-rigid motions. Secondly, unlike previous spatio-temporal graph optimization based approaches~\cite{dondera2014interactive,price2009livecut,wang2005interactive,badrinarayanan2010label,ramakanth2014seamseg,Tsai_CVPR_2016}, our method handles each frame independently. Therefore, a slight mis-segmentation in the current frame does not have induce critical effects on the following frames. The performance gap between the first and last image of the sequences (see \Fref{fig:graph}) and the stability in \Tref{tb:davis} highlight the robustness of our system to the drifting effect problem. Nonetheless, however, background objects of the same class and with an appearance similar to the target object are highly distractive in our system. Furthermore, the small size of the network output shows difficulty in handling thin objects.


\textbf{\textit{SegTrack}} \cite{FliICCV2013} To validate the performance of the proposed method more thoroughly, we also conducted experiments on this dataset using the following additional baseline algorithms: FST: Fast Object Segmentation in Unconstrained Video \cite{papazoglou2013fast}, DAG: Video Object Segmentation through spatially accurate and temporally dense extraction of primary object regions \cite{zhang2013video}, TMF: Video segmentation by tracking many figure-ground segments \cite{li2013video}, and KEY: Key-segments for video object segmentation \cite{lee2011key}. \Tref{tb:segtrack} summarizes the results. Here, our method shows comparable results but does not always outperform the other approaches. This is mainly due to the fact that our network has not been trained on the same dataset. In addition, the low-quality videos have caused the confusion for our network to distinguish the target object from the background.

\begin{table}[t]
	\centering
	\begin{tabular}[t]{p{1.2cm}|p{6mm}p{6mm}p{6mm}p{6mm}p{6mm}p{6mm}}   
		
		\toprule[1.5pt]  \\[-3.4ex]
		&   \scriptsize{TD}
		&   \scriptsize{AlexNet}
		&   \scriptsize{CN24}
		&   \scriptsize{FCN}
		&   \scriptsize{PLM}
		&   \scriptsize{PLM$_p$}
		\\[-0.5ex]\midrule[1.5pt]  \\[-3.2ex]
		
		
		\parbox{0.5cm}{\centering \scriptsize{campus}}
		&\scriptsize{10.7}&\scriptsize{42.1}&\scriptsize{36.4}&\scriptsize{\textcolor{blue}{\textbf{10.3}}}&\scriptsize{11.3}&\scriptsize{\textcolor{red}{\textbf{9.89}}}
		\\ [-0.5ex]\cmidrule{1-7} \\[-3.2ex]
		\parbox{0.5cm}{\centering\scriptsize{mountain}} 
		&\scriptsize{14.0}&\scriptsize{41.1}&\scriptsize{21.1}&\scriptsize{\textcolor{red}{\textbf{6.34}}}&\scriptsize{11.4}&\scriptsize{\textcolor{blue}{\textbf{10.4}}}
		\\ [-0.5ex]\cmidrule{1-7}\\[-3.2ex]
		
		\parbox{0.5cm}{\centering \scriptsize{city}}
		&\scriptsize{12.6}&\scriptsize{29.1}&\scriptsize{28.0}&\scriptsize{\textcolor{red}{\textbf{9.77}}}&\scriptsize{11.95}&\scriptsize{\textcolor{blue}{\textbf{11.5}}}		\\[-0.5ex]  \midrule[1.5pt]\cmidrule{1-7}\\[-3.2ex]
		
		\parbox{0.5cm}{\centering \scriptsize{Average}}
		&\scriptsize{11.8}&\scriptsize{39.0}&\scriptsize{36.7}&\scriptsize{\textcolor{red}{\textbf{9.45}}}&\scriptsize{11.5}&\scriptsize{\textcolor{blue}{\textbf{10.6}}}
		\\[-0.5ex]
		
		\bottomrule[2pt]  
	\end{tabular}   
	\vspace{1.5mm}
	\caption{Error rate (lower is better) evaluation on the \textit{Thermal-Road} \cite{yoon2016thermal} benchmark. \color{red}{\textbf{Red}}\color{black}{: best} , \color{blue}{\textbf{blue}}\color{black}{: second best.}}
	
	\label{tb:thermal}
	\vspace{-3.4mm}
\end{table}

\textbf{\textit{JumpCut}}\ \  In order to validate the effectiveness of our model on non-successive scenarios, we use 22 medium resolution videos published by Fan~\etal~\cite{fan2015jumpcut}. We compare our algorithm to the following methods: RB: RotoBrush tool from Adobe AfterEffect \cite{bai2009video} based on the SnapCut, and DA: Discontinuity-aware video object cutout \cite{zhong2012discontinuity}. We measured the performance using the same error metric described in~\cite{fan2015jumpcut}. Thus, we investigate the transferred mask from the i$^{th}$ key frame (i= 0, 16,..,96) to the (i+d)$^{th}$ frame. To do this, we fine-tune the network at the key frame and propagate mask skipping for 16 frames (d=16). We then calculate the average error score in each sequence according to the following equation:

\vspace{-4.2mm}
\begin{equation} \hspace{-3mm}
\label{error}
Err=\frac{100}{N_i}\sum_{i}^{} \frac{\#\ \textrm{of\ mislabeled pixels at}\ (i+d)^{th}\ \textrm{frame}}{\#\ \textrm{of foreground pixels at}\ (i+d)^{th}\ \textrm{frame}}, 
\end{equation}

where $N_i$ denotes the number of key frames. Overall, the proposed method outperforms all other methods, even without the active contour refinement process introduced in an earlier study~\cite{fan2015jumpcut}.

\textbf{\textit{Thermal-Road}}\ \  To demonstrate the applicability of our network to region based tasks such as road tracking, we present an evaluation of the thermal-infrared based road scene data from Yoon~\etal \cite{yoon2016thermal}, which contains three different scenarios for a total of approximately 6000 manually annotated images. To generate query and search data on this benchmark, we use the full frame (instead of bounding boxes), as the road occupies nearly the entire image. We conduct a comparison with one hand-crafted feature based method (TD: Thermal-infrared based drivable region detection \cite{yoon2016thermal}) and three different CNN based approaches (AlexNet \cite{krizhevsky2012imagenet}, CN24 \cite{brust2015convolutional}, and FCN \cite{long2015fully}). We follow the same error rate metric with \cite{yoon2016thermal} calculated by:\vspace{-1.5mm}
\vspace{-0mm}
\begin{equation} \hspace{-1mm}
\label{error}
ErrorRate=\frac{N_{FP}+N_{FN}}{N_{P}+N_{N}}\times 100, 
\vspace{-1mm}
\end{equation}

where $N_{FP}$, $N_{FN}$, $N_{P}$, and $N_{N}$ are the number of pixels which are respectively associated with incorrectly detected drivable and non-drivable region, and their ground-truth. As summarized in \Tref{tb:thermal}, our method outperforms the state-of-the-art hand-crafted feature based approach (TD), and performs second best overall, closely following the FCN approach. Note that, however, we use only an initial frame for fine-tuning our network pre-trained with color images, whereas FCN is fully supervised with the Thermal-Road datasets. From these experiments, we can validate that the proposed network is transferable to a different domain or task using only a single frame. The results also demonstrate that the proposed network is readily trainable with a simple fine-tuning step.

\vspace{-2mm}
\section{Conclusion}\label{sec:con}
\vspace{-1mm}

In this paper, we proposed a deep learning based video object segmentation algorithm. Our network is composed of encoding and decoding models which are suitable for pixel-level object matching. Two-stage training allows our network to handle appearance variations while also preventing over-fitting problem. We extensively evaluated our method on three widely used benchmark datasets. The obtained results demonstrate that our method performs better than previous related methods in terms of accuracy, speed, and stability. We also verified the importance of using multilayer features and a \textit{compression} technique to make the network compact while maintaining its object representation capability. Finally, we proved the transferability of our network to different domains using the thermal infrared database.
\section*{Acknowledgement}
\vspace{-2mm}
This work was supported by the Technology Innovation Program(No. 2017-10069072) funded By the Ministry of Trade, Industry \& Energy (MOTIE, Korea).

\clearpage
{\small
	\bibliographystyle{ieee}
	\bibliography{pix_matching}
}

\end{document}